\documentclass[a4paper,fleqn]{cas-sc}

\usepackage[authoryear,longnamesfirst]{natbib}

\usepackage{algorithmic}
\usepackage{algorithm} 
\usepackage{enumitem}
\usepackage{multirow}
\usepackage{bm}
\usepackage{soul}

\usepackage{amsthm}
\newtheorem{definition}{Definition}

\begin{document}
\let\WriteBookmarks\relax
\def\floatpagepagefraction{1}
\def\textpagefraction{.001}

\shorttitle{BAED: a New Paradigm for Few-shot Graph Learning with Explanation in the Loop}

\shortauthors{Chao Chen, Xujia Li et~al.}

\title [mode = title]{BAED: a New Paradigm for Few-shot Graph Learning with Explanation in the Loop}

\affiliation[1]{organization={Harbin Institute of Technology (Shenzhen)},
state={Guangdong},
country={China}}
\affiliation[2]{organization={The Hong Kong University of Science and Technology},
state={Hong Kong SAR},
country={China}}
\affiliation[3]{organization={Fuzhou University},
state={Fujian},
country={China}}
\affiliation[4]{organization={The Hong Kong University of Science and Technology (Guangzhou)},
state={Guangdong},
country={China}}

%

\author[1]{Chao Chen}
\ead{cha01nbox@gmail.com}
\fnmark[1]

\author[2]{Xujia Li}
\fnmark[1]

\author[3]{Dongsheng Hong}
\author[3]{Shanshan Lin}
\author[3]{Xiangwen Liao}
\author[1]{Chuanyi Liu}
\cortext[1]{Corresponding author}
\cormark[1]

\author[2,4]{Lei Chen}

\fntext[fn1]{Co-first authors}

\begin{abstract}
The challenges of training and inference in few-shot environments persist in the area of graph representation learning. The quality and quantity of labels are often insufficient due to the extensive expert knowledge required to annotate graph data. In this context, Few-Shot Graph Learning (FSGL) approaches have been developed over the years. Through sophisticated neural 
\textcolor{black}{architectures} 
and customized training pipelines, these approaches enhance model adaptability to new label distributions. However, compromises in \textcolor{black}{the model's} robustness and interpretability can result in overfitting to noise in labeled data and 
\textcolor{black}{degraded} 
performance. This paper introduces the first explanation-in-the-loop framework for the FSGL problem, called BAED. 
We novelly employ the belief propagation algorithm to facilitate label augmentation on graphs. Then, leveraging an auxiliary graph neural network and the gradient backpropagation method, our framework effectively extracts 
explanatory \textcolor{black}{subgraphs} surrounding target \textcolor{black}{nodes}. 
The final \textcolor{black}{predictions are} 
based on these informative subgraphs while mitigating the influence of redundant information from neighboring nodes. Extensive experiments on seven benchmark datasets demonstrate superior prediction accuracy, training efficiency, and explanation quality of BAED. As a pioneer, this work highlights the potential of the explanation-based research paradigm in FSGL.
\end{abstract}



\maketitle

\section{Introduction}

Graphs can clearly represent entities and their relatedness in the real world, such as social networks, transactions, and knowledge graphs. Training and inference in few-shot environments \textcolor{black}{are} primary but challenging \textcolor{black}{scenarios} for graph representation learning, represented by graph neural networks (GNNs) \citep{song2023comprehensive}, because acquiring labels for graph data often requires extensive expert knowledge, such as understanding of chemical molecular structures or social fraud patterns \citep{jin2017predicting, grover2016node2vec}, which makes it difficult to apply crowd-sourcing solutions that work well for image and text data. 

\textcolor{black}{In this context,}
Few-Shot Graph Learning (FSGL) has been developed for years. Most methods fall into two categories \citep{zhang2022few}. The optimization-based methods involve modifying the training pipelines, such as using meta-learning or transfer learning, to enhance model ability in quickly adapting to new label distribution with few-shot samples \citep{ding2021few, lan2020node}. Secondly, the metric-based methods summarize the limited labeled node information into corresponding class features, completing classification by calculating the feature similarity between the target node and the classes \citep{snell2017prototypical, yao2020graph}. 
However, both approaches still 
\textcolor{black}{exhibit} 
shortcomings in aspects of effectiveness, efficiency, generalizability, and explainability:

\begin{figure}[tp]
\centering
\includegraphics[width=3.3 in]{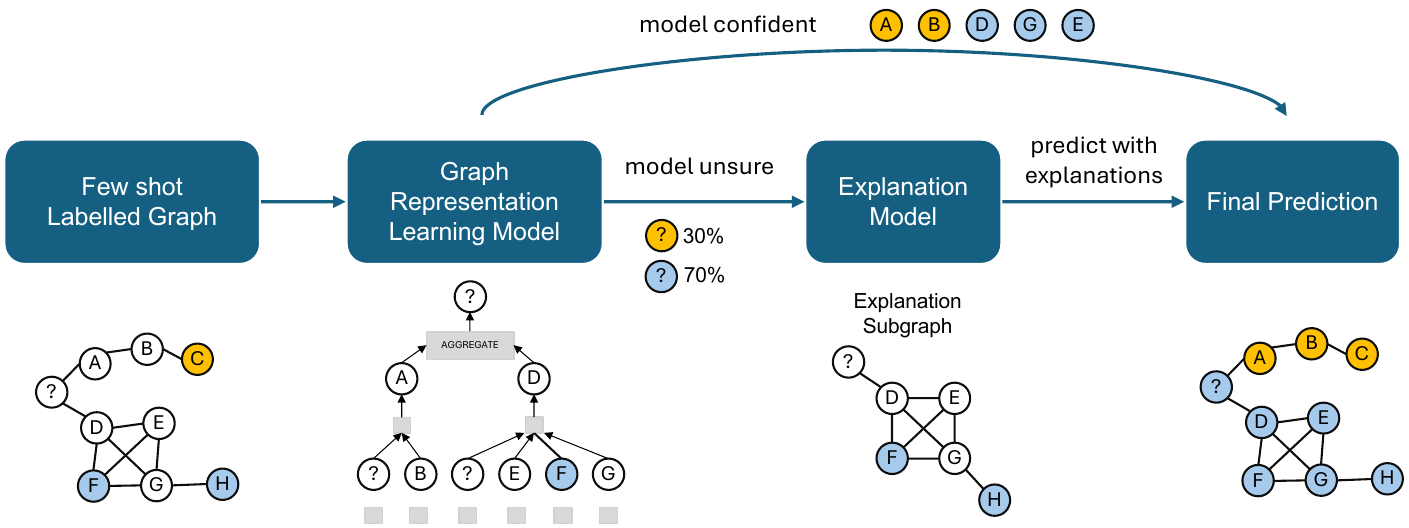}
\caption{Explanation in the loop: a new paradigm for FSGL}
\label{Fig1}
\end{figure}

First, from the aspect of \textbf{data quality}, the unstable labels with noise and the scarcity of node features still limit the model \textbf{effectiveness}. A small amount of noise is influential enough to push the few-shot label distribution away from the ground truth. Although optimization-based approaches can endow the model with the ability to quickly adapt to new classes, this larger flexibility often comes at the cost of reduced robustness facing noisy data \citep{cai2024semi}, resulting in blind obedience to new information and the catastrophic forgetting of previously learned knowledge \citep{carta2022catastrophic}. Furthermore, all current FSGL approaches assume the availability of node features for those unlabeled nodes. However, a feature-agnostic 
\textcolor{black}{scenario} 
is more common in practical scenarios and presents greater challenges. For instance, in social networks or e-commerce platforms, increasing demands for privacy protection means that only a small number of labeled nodes possess complete node features, while for the remaining nodes, only the surrounding graph structure can be accessed. Similar situations where only graph structure is available without labels or any features arise in cold-start, blockchain, and federated learning problems. In such cases, the metric-based methods fail to map the relation between features and labels, causing a deviated decision boundary \citep{song2022learning, yuan2023alex}.

Second, most existing FSGL methods adopt highly customized but inefficient designs, including intricate network architectures \citep{yao2020graph}, elaborate training schemes \citep{tan2022graph, lan2020node}, or additional training objectives \citep{ding2020graph,  zhang2022adapting}, which results in inconsistent \textbf{adaptability} of these methods to different data distributions in various applications. 
\textcolor{black}{Customizing the backbone GNNs} 
also introduces an increased model complexity with a larger number of trainable parameters, hindering training \textbf{efficiency} \citep{yu2022hybrid, li2024simple}. 
\textcolor{black}{Thus, there is an urgent need for a more efficient and universal research paradigm in FSGL.}

Third, \textbf{interpretability} is 
\textcolor{black}{particularly} 
more critical in few-shot scenarios, because only through explanations about predictions can humans determine whether the model has captured the correct features from the few-shot labels, excluding those false improvements caused by overfitting during training \citep{ju2024survey}. However, it is challenging to adapt current GNN explanation methods to few-shot scenarios \citep{zhang2022few}. Without abundant labels, post-hoc explanation methods cannot build a mapping between features and labels to explain the model predictions based on node features \citep{LIME}. Similarly, methods that use subgraphs as explanations are not feasible: Generation-based approaches require a ton of examples with labels to guide the subgraph generation \citep{yuan2020xgnn}; Extraction-based methods, which determine important edges and extract subgraphs by executing gradient backpropagation through GNN, also need a substantial number of labels to prevent gradients from vanishing during backpropagation \citep{baldassarre2019explainability}. Moreover, the explanation-in-the-loop pipeline is promising to address the effectiveness bottleneck of FSGL, because the high-quality explanations can help to eliminate the interfering noise in few-shot labels, as illustrated in \textcolor{black}{Figure.}~\ref{Fig1}.

In this context, we propose a new explanation-based framework for FSGL\textcolor{black}{,} named \textbf{BAED}, which offers superior 
\textcolor{black}{predictive} 
performance, training efficiency, and reliable explanations. The pipeline 
\textcolor{black}{consists of four main steps.} 

\textbf{(1) ``B'': BP-based Label Augmentation}. We incorporate the Belief Propagation (BP) algorithm as a key to resolving the first challenge related to the data quality in FSGL. 
\textcolor{black}{BP is a well-established algorithm for probabilistic inference, which estimates marginal probabilities in complex probabilistic graphs \citep{koller2009probabilistic}.}
It is commonly used in computer vision for image segmentation and in natural language processing for syntactic analysis. In this module, we utilize BP as a data augmentation process. The few-shot labels are transformed into prior probabilities and propagated across the graph. Through the iterative message-updating steps in BP, this process generates sufficient posterior probabilities as pseudo-labels, addressing the issue of label scarcity. This BP-based data augmentation allows each node to perceive global information over a larger range and mitigate the influence of 
\textcolor{black}{local noise} 
In contrast, conventional GNNs are unable to perceive information that exceeds their aggregation hops \citep{garg2020generalization}, so they are vulnerable to the 
\textcolor{black}{local noise} 
on sparse graphs \citep{li2023diga}. Moreover, this \textbf{augmentation of data quality} does not require the involvement of any node features and customized parameters in model architecture, making it \textbf{more universal} for various datasets. 

\textbf{(2) ``A'': Auxiliary GNN Training}. Unlike existing methods that directly use a GNN for final node classification, the auxiliary GNN in BAED aims to identify subgraphs that contain critical information relevant to the class of target nodes. Through mapping from priors to posteriors, this module learns how labels influence each other in specific topological structures. This subgraph-level estimation is less susceptible to the noises in labels compared to precise fitting to those few-shot labels. For \textbf{better training efficiency}, instead of consuming long feature vectors at the input layer, the auxiliary GNN takes prior probability vectors as input with 
\textcolor{black}{a much shorter length} 
greatly reducing the trainable parameters in the network.

\textbf{(3) ``E'': Explanatory Subgraph Extraction}. 
\textcolor{black}{The auxiliary GNN} 
trained on the augmented graph can obtain distinguishable gradients through backpropagation, making BAED a compatible host for all gradient-based explanation approaches. We extract the top-N relevant neighbors of the target node to construct an explanatory subgraph based on the magnitude of the gradients on their connecting edges. By analyzing typical topological patterns and the features of labeled nodes within the subgraph, \textbf{generalizable interpretability} is achieved in few-shot scenarios.

\textbf{(4) ``D'': Decision Making with Explanation in the Loop}. Essentially, previous modules can be treated as a diffusing and then refining operation on labeled information, ensuring that the explanatory subgraph encompasses the decisive factors necessary for classifying the target nodes while
\textcolor{black}{eliminating nearby redundant noise} 
We invoke the BP algorithm again on the subgraphs to make the final prediction, eventually \textbf{addressing the model effectiveness limitation} with this explanation-in-the-loop manner.

We summarize the core contributions of this work:

1. We propose BAED, an end-to-end framework, with the novel BP-based label augmentation technique and the explanation-guided prediction manner to solve the existing bottlenecks of FSGL models.

2. BAED enables the most gradient-based subgraph explanation methods in few-shot scenarios. It illuminates the research on the interpretability of FSGL and establishes a foundation with high compatibility for further exploration.

3. We conduct extensive experiments with twenty related methods on seven benchmark datasets, demonstrating that BAED significantly outperforms state-of-the-art baselines in terms of effectiveness, efficiency, and explainability. 

\section{Related Work}

\subsection{Few-Shot Graph Learning}

Few-Shot Graph Learning (FSGL) has been developed over many years. 
\textcolor{black}{Related work on FSGL for node classification can be broadly divided into metric-based and optimization-based methods \cite{song2023comprehensive}.}
Metric-based methods, such as ProNet \citep{snell2017prototypical}, 
\textcolor{black}{learn node embeddings} 
through an encoder and then averaging the embeddings of labeled nodes within each class to create a representative class embedding. Classification is achieved by comparing the distance between the query node and each class embedding. Optimization-based methods are primarily based on concepts like model-agnostic meta-learning or transfer learning. For example, Meta-GNN is trained with several sampled meta-training tasks, allowing the model to quickly adapt to gradient update patterns in few-shot scenarios \citep{metagnn}. 

Although these two approaches have achieved certain successes in various settings, unresolved issues persist regarding node classification performance, model compatibility, and interpretability in FSGL.

\subsection{Explainability of Graph Neural Networks}

The black-box nature of deep neural networks limits the application of GNNs in many critical decision-making scenarios \citep{kakkad2023survey}, such as fraud detection and molecule analysis, 
\textcolor{black}{where the need for explainability often outweighs classification performance \citep{li2023diga}.}
Existing explainable GNNs can be categorized into two granularity of explanations \citep{yuan2022explainability}. The first category provides explanations for individual nodes or edges, including node importance analysis \citep{pope2019explainability}, similarity identification \citep{feng2023degree}, and edge pattern analysis \citep{schnake2021higher}. The second level involves explanations at the subgraph or graph level, allowing for a broader discussion of topological structures and neighborhood characteristics \citep{kakkad2023survey}. Perturbation-based methods, such as GNNExplainer \citep{gnnexp}, identify important subgraphs by perturbing the input data \citep{schlichtkrull2020interpreting, pgexp}. Generation-based methods, represented by XGNN \citep{yuan2020xgnn}, utilize reinforcement learning to maximize the probability of inference on a generated subgraph to a particular class. 
\textcolor{black}{A third line of work} 
is based on the gradient back-propagation. The gradients represent the rate of change and can indicate the sensitivity of outputs to the variations in inputs, thereby highlighting key edges to construct an explanatory subgraph for the prediction \citep{baldassarre2019explainability}. 

\section{Problem Definition}

The graph $\mathcal{G}$ consists of nodes $\mathcal{V}=\{v_i\,|\,i=1,2,...,n\}$ and their connected edges $(v_i,v_j)\in\mathcal{R}$. The node feature vector $\bm{x}_i$ represents the features of each node $v_i$. which constitute the graph's feature matrix $X$. $y_i$ denotes the label of node $v_i$, and each label takes values in the class set $\mathcal{C}= \{1,2,...\}$, where $|\mathcal{C}|$ is the number of classes. The adjacency matrix, $A=\{0,1\}^{n\times n}$, describe the topology of the graph. $A_{i,j}=1$ represent that an edge exists between node $v_i$ and node $v_j$, while $A_{i,j}=0$ means no relation between them. For a clear presentation, we summarize the major notations in Table~\ref{tab_not}, where vectors are represented in bold font, and the notation $\bm{\phi}_i(c_i)$ denotes the $c_i$-th value in the vector $\bm{\phi}_i$.

\begin{table}[htbp]
\small
    \renewcommand{\arraystretch}{1.2}
    \caption{Major notations}
    \label{tab_not}
    \centering
    \scalebox{0.9}{
\begin{tabular}{@{}cl@{}}
\toprule
Notation        & Description                                \\ \midrule
$\bm{x}_{i}$    & Feature vector of a node $v_i$             \\
$c_i$              & A variable representing one potential class of node $v_i$, $c_i\in \mathcal{C}$ \\
$\bm{\hat{c}}_i$   & Class probability distribution vector of $v_i$ predicted by models              \\
$\bm{\phi}_i$   & Prior probability vector of node $v_i$     \\
$\bm{\phi}_i(c_i)$ & Prior probability of node $v_i$ belonging to one specific class $c_i$           \\
$\bm{b}_i$      & Posterior probability vector of node $v_i$ \\
$\psi(c_i,c_j)$    & Compatibility potential between node $v_i$ and $v_j$ in BP                      \\
$\mathcal{G}_i$ & Explanatory subgraph for target node $v_i$ \\
$\mathcal{N}_i$ & Set of one-hop neighbors of node $v_i$     \\
$y_i, \hat{y}_i$   & Ground truth and the predicted class of node $v_i$                              \\ \bottomrule
\end{tabular}
    }
\end{table}

\begin{definition} Node Classification Task in FSGL \label{Definition 1}\\
Given the graph $\mathcal{G}=(X,A)$ with partially labeled nodes $\mathcal{V}_{labeled}\subset \mathcal{V}$, the task is to predict the class $\hat{y}$ of unlabeled nodes $\mathcal{V}_{target}= \mathcal{V} \backslash \mathcal{V}_{labeled}$, where only the feature vectors of labeled nodes are available. And $r=|\mathcal{V}_{labeled}|/|\mathcal{V}|$ denotes the labeling ratio of the specific graph dataset.
\end{definition}

\section{Methodology}

\begin{figure*}[t]
\centering
\includegraphics[width=6.2 in]{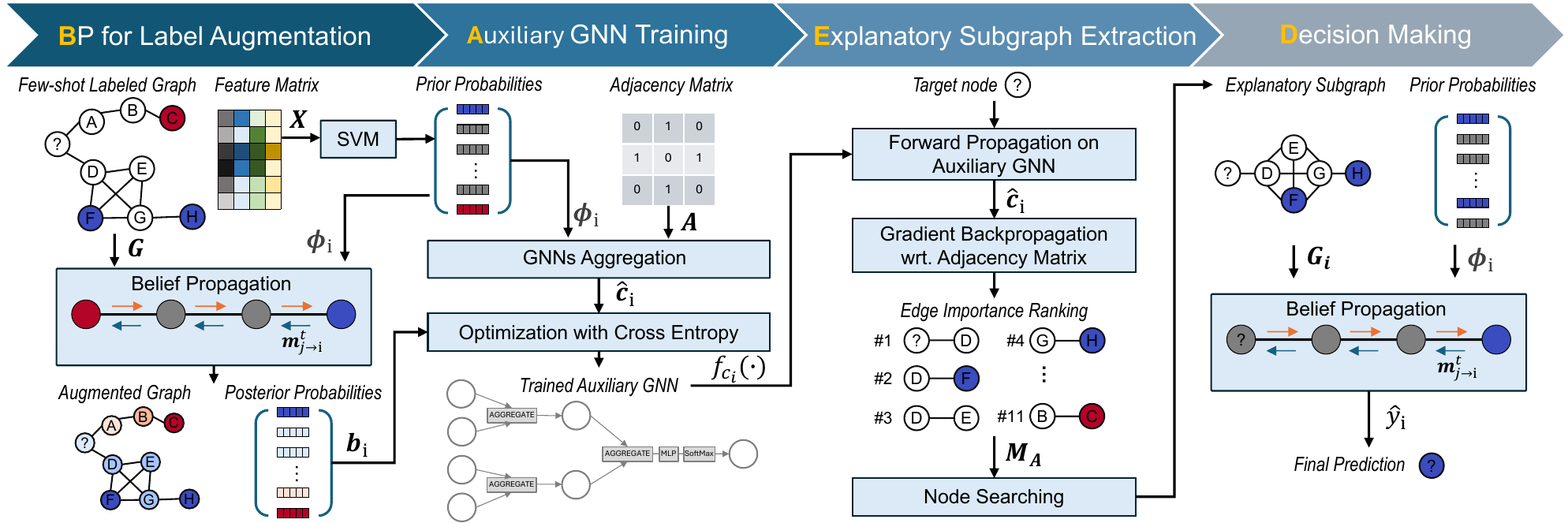}
\caption{BAED: the explanation-based FSGL pipeline}
\label{Fig2}
\end{figure*}

As illustrated in \textcolor{black}{Figure.}~\ref{Fig2}, BAED is an end-to-end pipeline containing four modules from label augmentation to auxiliary GNNs training, subgraph extraction, and final prediction. Correspondingly, all innovations are clarified in detail in the following four subsections.

\subsection{Belief Propagation for Label Augmentation}

Label augmentation module is employed to address the insufficient label quantity and quality in the FSGL problem. Additionally, this module mitigates the limitations of GNNs in perceiving distant graph structures out of their aggregating hops. In BAED, the label augmentation is formulated as a process in Markov Random Field (MRF) and utilizes the Belief Propagation (BP) algorithm to propagate the class information from labeled nodes, referred to as prior probabilities, to a broader range of unlabeled nodes, referred to as posterior probabilities. The posterior probabilities will function as the fitting objectives in the subsequent auxiliary GNN training.

Based on the assumption of MRF, the probability of each node's label is influenced solely by its neighbors and is independent of other non-adjacent nodes. Thus, the joint probability $P(\mathcal{V})$ of this MRF can be factorized with node information as follows:
\begin{equation}
\label{eq1}
P(\mathcal{V}) \cong \prod_{v_i\in \mathcal{V}} \bm{\phi}_i(c_i) \prod_{(v_i,v_j)\in \mathcal{R}}\psi(c_i, c_j)\,,
\end{equation}
where we use $\cong$ to denote equality after normalization. $P(\mathcal{V})$ describes the joint probability of a specific case that each node $v_i\in \mathcal{V}$ on the graph has a respective class $c_i$. The term $\bm{\phi}_i(c_i)$ represents the prior probability of node $v_i$ belonging to the class $c_i$. The prior probability is often initialized based on prior knowledge about the nodes, which can be raw features, degrees, labels, etc \citep{Rayana2015}. We employ a support vector machine (SVM) to 
\textcolor{black}{map} 
the raw features $\bm{x}_i$ of labeled nodes $v_i \in \mathcal{V}_{\text{labeled}}$ onto their prior probability vectors $\bm{\phi}_i$, of which the length is the total number of classes $|\mathcal{C}|$. For instance, the second element in vector $\bm{\phi}_i$ indicates the probability of node $v_i$ being labeled as the second class $c_i=2$. As a lightweight machine learning model, SVM only requires a small amount of data to ensure that the prior probabilities embed the node features and their high-dimensional associations with the labels. Concurrently, we initialize the prior probability vector of unlabeled nodes using the normal distribution, as a starting point of the BP algorithm. 

In the equation \ref{eq1}, $\psi(c_i, c_j)$ denotes the compatibility potential matrix between node $v_i$ and node $v_j$. This potential quantifies the likelihood of two connected nodes taking the label $c_i$ and $c_j$ jointly. The rationale behind defining compatibility potentials is to enhance message passing in BP through homogeneous edges when two nodes share the same class \citep{Rayana2015}. Conversely, a penalty is imposed on the message passing between nodes belonging to different classes. The compatibility matrix utilized by BAED is presented as follows:
\begin{equation}
\label{eq_comp}
\psi(c_i, c_j)=\left\{
\begin{aligned}
\begin{split}
&\epsilon \quad when \ c_i = c_j \\
&(1-\epsilon)\,/\,(|\mathcal{C}|-1) \quad  when \ c_i \neq c_j\,,
\end{split}
\end{aligned}
\right.
\end{equation}
where $\epsilon$ is the penalty hyperparameter. A higher penalty factor indicates that less information 
\textcolor{black}{is} 
transmitted through the edges connecting two different classes.

Based on the above definition of MRF, the BP process is propelled with continuous message passing through neighboring nodes. A single message passed between two nodes at the $t$-th iteration from node $v_j$ to node $v_i$ denoted as $\bm{m}_{j \to i}^t$. The updating rule of messages during BP process is calculated as follows:
\begin{equation}
\label{eq_msg}
\bm{m}_{j \to i}^{t+1}(c_i)\cong\sum_{c_j=1}^{|\mathcal{C}|}
\left[\psi(c_i, c_j)\bm{\phi}_j(c_j) \prod_{v_k\in {\mathcal{N}_j}\setminus \{v_i\}} \bm{m}_{k\to j}^t(c_j)\right]\,,
\end{equation}
with the normalization, the message vector adheres to a probability distribution. The initialization of $\bm{m}_{k\to j}^{t=0}$ follows a uniform distribution \citep{bp1,GE-L}. Consequently, after running one update of the BP algorithm, the stopping flag is examined as follows:
\begin{equation}
\label{eq_stop}
\eta>\frac{1}{2|\mathcal{R}|}\sum_{(v_i,v_j)\in\mathcal{R}} \sum_{c_i=1}^{|\mathcal{C}|} |\bm{m}_{j \to i}^{t+1}(c_i)-\bm{m}_{j \to i}^{t}(c_i)|\,.
\end{equation}
We use Manhattan distance to compute the update amplitude between consequent iterations. If the graph is undirected, this amplitude should be counted twice for both directions. The hyperparameter $\eta$ acts as the stopping signal. Once the average update on each edge in two consecutive iterations falls within this threshold, 
BP is terminated at $t=T$. 

The posterior probability, the marginal class probability distribution of node $v_i$, is then calculated as follows:
\begin{equation}
\label{eq_post}
    \bm{b}_i \cong \, \bm{\phi}_i\prod_{v_j\in {\mathcal{N}_i}} \bm{m}_{j\to i}^{T}\,.
\end{equation}
Compared to the prior probabilities, which are based solely on node-wise features, the posterior probabilities better estimate all node labels because they encompass a broader range of topological structures and incorporate information from few-shot labels via BP. Therefore, the posterior probabilities serve as a more robust objective for the GNN model to fit. Algorithm~\ref{algo1} outlines the complete procedure of label augmentation with BP in FSGL.

\begin{algorithm}[htbp]
	\caption{Label Augmentation with BP}
    \label{algo1}
	\begin{algorithmic}[1]
	\REQUIRE Graph $\mathcal{G}$, Node features and Labels $\{(\bm{x}_i, y_i)\,|\,\forall v_i\in \mathcal{V}_{labeled}\}$
		\STATE $\forall v_i\in \mathcal{V}_{labeled}$, initialize priors $\bm{\phi}_i\leftarrow SVM(\bm{x}_i,y_i)$
        \STATE $\forall v_i\in \mathcal{V}_{target}$, initialize uniform priors $\bm{\phi}_i(c_i) \leftarrow \frac{1}{|\mathcal{C}|} $
        \STATE $\forall (v_i,v_j) \in \mathcal{R}$, initialize uniform messages $\bm{m}_{j\to i}^{t=0}(c_i) \leftarrow \frac{1}{|\mathcal{C}|} $
		\STATE \textbf{do}:
			\STATE \quad $\forall (v_i,v_j) \in \mathcal{R}$, update messages based on Eq. (\ref{eq_msg})
            \STATE \quad $t\leftarrow t+1$
		\STATE \textbf{until} $\eta>\frac{1}{2|\mathcal{R}|}\sum_{(v_i,v_j)\in \mathcal{R}}\ \|\bm{m}_{j\to i}^{t+1}-\bm{m}_{j\to i}^{t}\|_{p=1}$ 
			
        \STATE $\forall v_i\in\mathcal{V}$, compute posterior probabilities $\bm{b}_i$ based on Eq. (\ref{eq_post})
    \RETURN $\{\bm{b}_i\,|\,i=1,2,...,n\}$
	\end{algorithmic}
\end{algorithm}

\subsection{Auxiliary GNN Training}

\textcolor{black}{Directly using GNNs for prediction on graphs with extremely few labels limits accuracy  due to the lack of effective training samples.}
Meanwhile, most modifications to the training schema or direct summarization from few-shot labels in existing FSGL models are susceptible because even minor disturbances in labels can cause the model's learned distribution to deviate significantly from the true distribution. Instead, BAED emphasizes that predictions should be made based on explanatory subgraphs containing the most critical information for decision-making while eliminating redundant noise nearby. Ablation studies in section \ref{abla} and related tests about explanatory subgraphs in Table~\ref{Tab_Subgraph} support our motivations of explanation-guided learning. 

In this module, we train an auxiliary GNN specifically for extracting an explanatory subgraph as a process of information refinement and noise reduction, enabling the identification of key subgraphs that assist the final prediction module. The training target is to align the vector of GNN outputs $\bm{\hat{c}}_i$ with the posterior probabilities $\bm{b}_i$. The input of GNN is prior probabilities $\bm{\phi}_i$ and the adjacency matrix $A$. With this objective, the GNN can capture the high-dimensional projection from the node-level information in the priors to the topological patterns condensed in the posteriors. 
\textcolor{black}{This fitting process benefits from sufficient training samples, since all nodes have prior and posterior probabilities after the label augmentation with BP.}

Rather than introducing intricate training schemes or designing additional components, our framework emphasizes compatibility with arbitrary GNN backbones and arbitrary subgraph-based explanation methods, so that BAED can lay the groundwork for further exploration in explanation-based FSGL. We select \textcolor{black}{SAGE} \citep{hamilton2017inductive}, a well-established GNN, as an example. At the zeroth layer ($k=0$), the hidden embedding $\bm{h}_i^0$ is initialized with the prior probability:
\begin{equation}
\bm{h}_i^0 = \bm{\phi}_i\,, \,\forall v_i \in \mathcal{V}\,.
\end{equation}
After the $k$-th aggregation step in the network, the embedding can be computed from neighboring nodes of node $v_i$. 
\begin{equation}
\label{eq_gnn}
\begin{split}
& \bm{h}_{i}^k =  \sigma \left(W^k\cdot Mean(\{\bm{h}_{i}^{k-1} \cup  \mathcal{H}_{\mathcal{N}_i}^{k}\} \right) \\
& with \,\,\, \mathcal{H}_{\mathcal{N}_i}^{k} := \{{\bm{h}_{j}^{k-1}, \forall v_j \in \mathcal{N}_i}\}\,,
\end{split}
\end{equation}
where we take $Mean(\cdot)$ as the default aggregator function, $\sigma$ represents \textcolor{black}{the} ReLU activation function, $W$ denotes \textcolor{black}{the} trainable parameters of neural networks. The final hidden embedding $ \bm{h}_{i}^K$ is transformed into a multi-class classification probability $\hat{\bm{c}}_{i}$ through multiple layers of a Multi-Layer Perceptron (MLP) and a Softmax operation. Eventually, the loss function is defined as the cross-entropy between the outputs and the posterior probabilities:
\begin{equation}
\label{loss}
\mathcal{L}(\bm{b}_i, \, \hat{\bm{c}}_{i}) = CrossEntropy(\bm{b}_i,SoftMax(MLP(\bm{h}_{i}^K))\,.
\end{equation}

\subsection{Explanatory Subgraph Extraction and Method Interpretability}

With the trained auxiliary GNN, this module aims to extract a high-quality explanatory subgraph for each target node. Due to the inherent characteristics of GNNs, the output probability distribution can be backpropagated to the input adjacency matrix using gradients \citep{SM}. These gradient values can be interpreted as the significance of each edge concerning a particular prediction outcome \citep{wang2024gradient}. A subgraph can be constructed by ranking and selecting the top-N important edges, typically encompassing the decisive factors for classifying the target node. 

After the forward process through auxiliary GNN for each target node, we perform gradient backpropagation. \textcolor{black}{The saliency map $M_A^{(v_i,v_j)}$ illustrates the importance of edges, which is computed by taking the gradient of the output with respect to the adjacency matrix:
\begin{equation}
\label{eq_grad}
M_A^{(v_i,v_j)} = \left| \frac{\partial \mathcal{L}(\bm{b}_i,f_{c_i}(A, X))}{\partial A_{v_i,v_j}} \right|, \quad \forall (v_i, v_j) \in \mathcal{R} \,,
\end{equation}
where $f_{c_i}(\cdot)$ denotes the output of auxiliary GNN. Based on the saliency map, we initiate a search starting from the target node $v_i$. The most relevant neighbor $v_j^*$ is determined by the edge with the highest importance, \[ (v_i, v_j^*) = \underset{(v_i, v_j) \in \mathcal{R}}{\arg\max} \; M_A^{(v_i, v_j)}.\]} 
These two nodes are then combined into a node-set. Subsequently, we search for the next node connected to this set by the edge with the highest importance. The extraction continues until a total of $N$ nodes are selected to construct the explanatory subgraph. 

\begin{figure}[htbp]
\centering
\includegraphics[width=2.5 in]{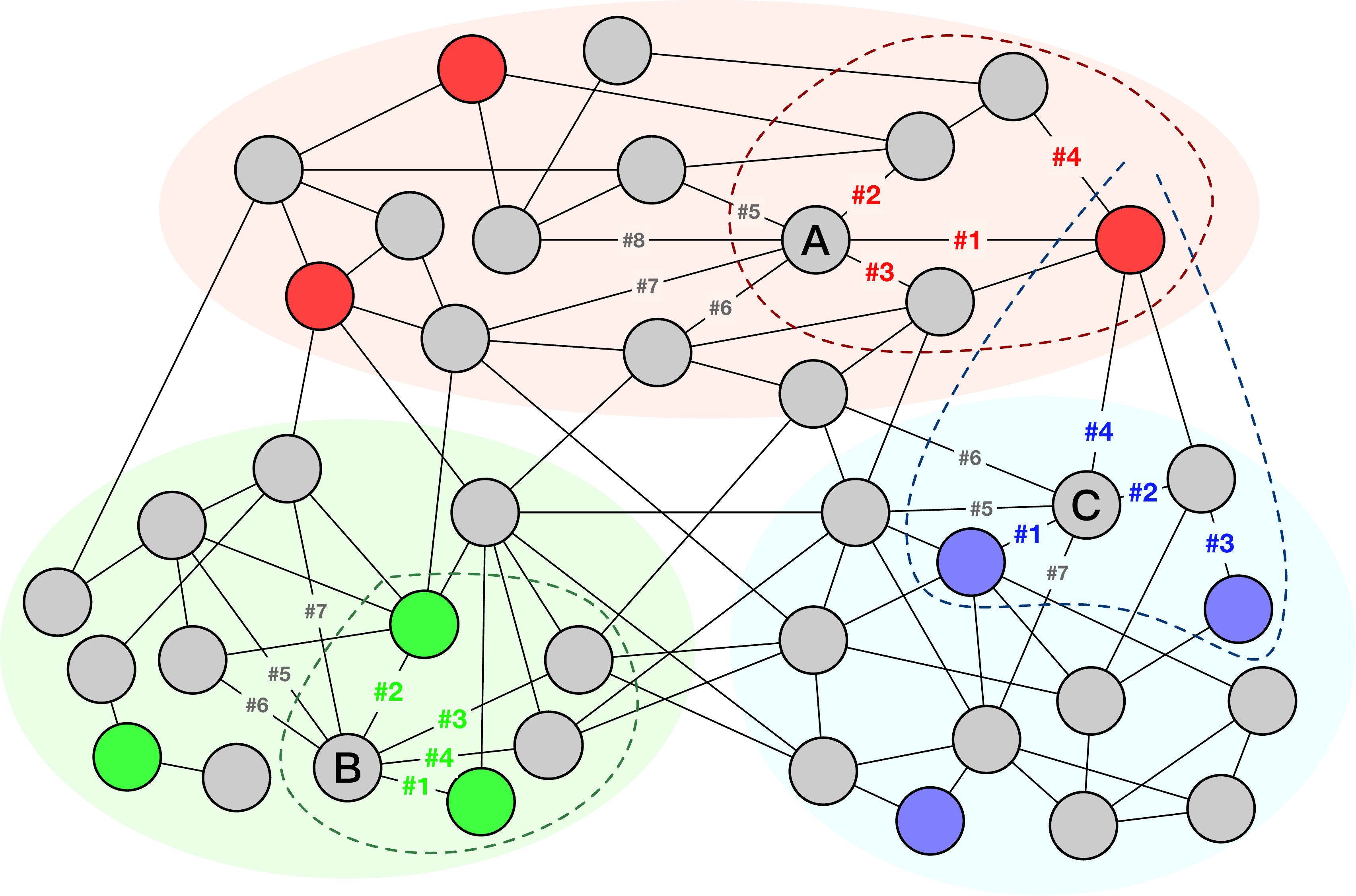}
\caption{\textcolor{black}{Explanatory subgraphs extracted using gradient backpropagation and edge ranking on the PubMed.}
\textcolor{black}{Colored backgrounds (red, green, blue) indicate ground-truth node labels; gray nodes are unlabeled. 
Dashed circles denote the explanatory subgraphs for node A, B, and C, with edge numbers indicating importance rankings.
Nodes A and B lie within homogeneous regions of same-label nodes, resulting in clean subgraphs without distracting connections. 
Node C, positioned near both red and blue nodes, yields a mixed but predominantly blue subgraph, leading to a correct prediction.
}
}
\label{Fig_Case}
\end{figure}

\textbf{\textcolor{black}{Case study.}}
\textcolor{black}{To better demonstrate the interpretability of BAED and to elucidate the decision basis of the model using explanatory subgraphs, we present the extracted explanatory subgraphs on the PubMed dataset. 
As illustrated in Figure~\ref{Fig_Case}, each target node's explanatory subgraph includes labeled information. The related edge ranking also confirms that through gradient back-propagation, \textbf{BAED can effectively identify the decisive factors behind each prediction.}}

BAED is able to accommodate any subgraph extraction algorithm based on gradient backpropagation, such as Integrated Gradients \citep{inpgrad}, and GNNExplainer \citep{gnnexp}. \textcolor{black}{To better lay the foundation for explainable FSGL, we compare the interpretability of seven subgraph-based explanation methods in the experiment section \ref{sec-exp}. And we present the formal definition of the method interpretability in FSGL as follows:}

\begin{definition} Method Interpretability in FSGL
\label{Definition 2}\\
\textcolor{black}{Faithfulness, as a metric that does not require interpretability-related labels or human evaluation, is widely employed in the subgraph-based explanation methods. Its essence lies in measuring the similarity between predictions made on the explanatory subgraph and those made on the original graph \citep{lv2023data}. Given the graph $\mathcal{G}$, and a target node $v_{i}\in \mathcal{V}_{target}$, extract a subgraph $\mathcal{G}_i \subset \mathcal{G}$, where the size limitation of the extracted subgraph is $N$. The faithfulness of $\mathcal{G}_i$ is computed with the symmetric KL-divergence between the class probability distribution $\hat{\bm{c}}_{\mathcal{G}}$ on the whole graph and the one on the explanatory subgraph $\hat{\bm{c}}_{\mathcal{G}_i}$: 
\begin{equation}
\label{eq:kl-div}
D_{KL}(\hat{\bm{c}}_{\mathcal{G}}||\hat{\bm{c}}_{\mathcal{G}_i})= \sum_{x=1}^{|\mathcal{C}|} \hat{\bm{c}}_{\mathcal{G}}(x)\log\frac{\hat{\bm{c}}_{\mathcal{G}}(x)}{  \hat{\bm{c}}_{\mathcal{G}_i}(x)}\,,
\end{equation}
where a small KL distance means a faithful explanatory subgraph concerning the prediction made on the original graph \citep{1992Explanation}.}
\end{definition}

To verify that the explanatory subgraph contains critical information for the final predictions, we conduct an analysis from a statistical perspective to evaluate the effectiveness of the explanatory subgraph in Table~\ref{Tab_Subgraph}. As shown in the first column about the Cora dataset, when the labeling ratio is $1\%$, the probability of extracting a subgraph containing a labeled node through random walks is quite low at $2.5\%$. However, BAED significantly increases this probability across all datasets. This indicates that the subgraph extraction module can effectively extract meaningful subgraphs with labeled information, thereby compelling the model to focus on critical information for prediction.

\begin{table}[htbp]\footnotesize
\centering
\renewcommand{\arraystretch}{1.3}
\setlength\tabcolsep{2.5 pt}
\caption{Proportion of Subgraphs with Labels \textbf{$\uparrow$}}
\begin{tabular}{@{}l||c|c|c|c|c|c|c@{}}
\toprule
         Datasets    & Cora   & Citeseer & PubMed & Wiki   & DBLP   & Wisconsin & CoauthorCS  \\ \midrule
Random walk & 2.5  & 3.6    & 2.2  & 4.2  & 2.8  & 4.9     & 4.4            \\
BAED        & 94.6 & 94.3   & 87.2 & 92.3 & 96.8 & 93.9    & 95.5        \\ \bottomrule
\end{tabular}
\label{Tab_Subgraph}
\end{table}

\subsection{Decision Making on Explanatory Subgraphs}

After pruning less relevant nodes from the graph, we apply the BP algorithm to the explanatory subgraph for the final prediction. The choice to utilize BP is out of two key considerations. First, the preceding three modules of BAED ensure that the core neighboring nodes are selected, thoroughly considering node features, topological structure, and label characteristics. BP can directly utilize prior probabilities for prediction, thereby preserving essential label information from being overly processed. Second, supervised learning methods are neither efficient nor feasible in the absence of appropriate labels for the explanatory subgraphs.

The final prediction is determined by the class with the highest probability in the posterior probabilities:

\begin{equation}
    \hat{y}_i = \mathop{argmax}_{c_i\in \mathcal{C}} \, \beta \,\bm{\phi}_i(c_i)\prod_{v_j\in {\mathcal{N}_{\mathcal{G}_i}}} \bm{m}_{j\to i}(c_i)\,,
\end{equation}
where $\mathcal{N}_{\mathcal{G}_i}$ is the neighborhoods of node $v_i$ on the explanatory subgraph $\mathcal{G}_i$ and $\beta$ denotes a normalization constant. The message updating and stopping criteria are in line with Eq. (\ref{eq_msg}) and Eq. (\ref{eq_stop}).

\subsection{Complexity Analysis of BAED}

The time complexity is $O(|\mathcal{V}||\mathcal{C}|^D)$ for the label augmentation module; $O(\prod_{k=1}^{K}S_k)$ for the auxiliary GNN training; $O(|\mathcal{R}|\log{N})$ for subgraph extraction; $O(N|\mathcal{C}|^D)$ for explanation-guided decision making, where $D$, $K$, $S_k$, $|\mathcal{R}|$, $N$ are the average degree of graph dataset, the number of GNN layers, sampling constants in \textcolor{black}{SAGE}, the number of edges, the number of nodes on explanatory subgraph. All modules maintain complexity within linear logarithmic order, ensuring the efficiency and scalability of the entire framework, verified in the efficiency test, shown in Table~\ref{tabEff}.

\section{Experiments}

In this section, we comprehensively evaluate our framework, including experiments on (1) few-shot node classification performance, (2) explainability, (3) training efficiency, (4) ablation studies on each module, (5) sensitivity analysis on different labeling ratios, (6) case studies, (7) convergence tests of BP. 



\subsection{Datasets}
We selected eight datasets encompassing various scenarios, such as citation networks, web pages, and co-authorship, to validate the broad applicability of our method. These datasets exhibit diverse characteristics in terms of feature numbers, class counts, average degree, and graph size, as listed in Table~\ref{tab:datasets}.

\begin{table}[b]
    \caption{Statistics of Datasets}
    \renewcommand{\arraystretch}{1.3}
    \setlength\tabcolsep{2.5 pt}
    \label{tab:datasets}
   \footnotesize
    \centering
    \begin{tabular}{@{}l|c|c|c|c|c@{}}
    \toprule
    \textbf{Datasets} & \textbf{\# Classes} & \textbf{\# Features} & \textbf{\# Edges} & \textbf{\# Nodes} & \textbf{Avg. Degree} \\ \midrule
    \textbf{Cora} \citep{Namata2016}       & 7  & 1,433 & 5,429   & 2,708  & 3.90  \\
    \textbf{Citeseer} \citep{bojchevski2017deep}   & 6  & 3,703 & 4,715   & 3,312  & 2.85  \\
    \textbf{PubMed} \citep{dernoncourt2017pubmed}     & 3  & 500   & 88,676  & 19,717 & 4.50  \\
    \textbf{Wiki}  \citep{wiki}      & 17 & 4,973 & 17,981  & 2,405  & 14.95 \\
    \textbf{DBLP}  \citep{Tang:08KDD}      & 3  & 1,639 & 105,734 & 17,716 & 11.94 \\
    \textbf{Wisconsin} \citep{pei2020geom}  & 5  & 1,703 & 515     & 251    & 4.10  \\
    \textbf{CoauthorCS} \citep{shchur2018pitfalls} & 15 & 6,805 & 163,788 & 18,333 & 17.87 \\ 
    \textcolor{black}{\textbf{CoauthorPhy}} \citep{shchur2018pitfalls} & \textcolor{black}{5} & \textcolor{black}{8,415} & \textcolor{black}{495,924} & \textcolor{black}{34,493} & \textcolor{black}{28.76} \\
    \bottomrule
    
    \end{tabular}
\end{table}

\begin{table*}[t]\footnotesize
\centering
\renewcommand{\arraystretch}{1.12}
\setlength\tabcolsep{2.9 pt}
\caption{Overall performance comparison on classification \textbf{$\uparrow$}}
\label{tabClassification}
\begin{tabular}{@{}l||cccc|ccc|ccc|cc|c@{}}
\toprule
\multirow{2}{*}{Datasets} &
  \multicolumn{4}{c|}{Traditional GNNs} &
  \multicolumn{3}{c|}{Advanced GNNs} &
  \multicolumn{3}{c|}{FSGL Models} &
  \multicolumn{2}{c|}{\textbf{BAED}} &
  \multirow{2}{*}{Improve.} \\ \cmidrule(lr){2-13}
 &
  \multicolumn{1}{c|}{\textcolor{black}{SAGE}} &
  \multicolumn{1}{c|}{GAT} &
  \multicolumn{1}{c|}{GIN} &
  GCN &
  \multicolumn{1}{c|}{SGC} &
  \multicolumn{1}{c|}{DLRGAE} &
  \multicolumn{1}{c|}{\textcolor{black}{HiD}} & 
  \multicolumn{1}{c|}{\textcolor{black}{GPN}} & 
  \multicolumn{1}{c|}{tsGCN} &
  DCI &
  \multicolumn{1}{c|}{\textcolor{black}{SAGE+IG}} &
  \textcolor{black}{SAGE+SM} &
   \\ \midrule
Cora       & 0.301  & 0.307 & 0.269  & 0.311* & 0.301  & 0.301 & \textcolor{black}{0.300} & \textcolor{black}{0.251} & 0.296  & 0.238  & 0.515          & \textbf{0.518} & 66.5\%  \\
Citeseer   & 0.202  & 0.192 & 0.215* & 0.199  & 0.189  & 0.189 & \textcolor{black}{0.204} & \textcolor{black}{0.199} & 0.211  & 0.199  & 0.769          & \textbf{0.787} & 266.1\% \\
PubMed     & 0.400  & 0.393 & 0.392  & 0.396  & 0.410  & 0.399 & \textcolor{black}{0.484} & \textcolor{black}{0.429} & 0.492* & 0.459  & 0.487          & \textbf{0.507} & 3.0\%   \\
Wiki       & 0.157 & 0.152 & 0.151  & 0.152  & 0.152  & 0.152 & \textcolor{black}{0.161*} & \textcolor{black}{0.158} & 0.152  & 0.139  & 0.172          & \textbf{0.176} & \textcolor{black}{9.3\%}  \\
DBLP       & 0.443  & 0.443 & 0.443  & 0.443  & 0.448  & 0.443 & \textcolor{black}{0.447} & \textcolor{black}{0.517*} & 0.442  & 0.449 & 0.690          & \textbf{0.698} & \textcolor{black}{24.6\%}  \\
Wisconsin  & 0.273  & 0.269 & 0.273  & 0.269  & 0.396  & 0.380 & \textcolor{black}{0.273} & \textcolor{black}{0.407} & 0.269  & 0.420* & \textbf{0.441} & 0.420          & 4.9\%   \\
CoauthorCS & 0.226  & 0.216 & 0.119  & 0.183  & 0.280* & 0.181 & \textcolor{black}{0.278} & \textcolor{black}{0.259} & 0.255  & 0.260  & \textbf{0.210} & 0.192          & -25.1\% \\ 
\textcolor{black}{CoauthorPhy} & \textcolor{black}{0.308} & \textcolor{black}{0.315} & \textcolor{black}{0.308} & \textcolor{black}{0.302} & \textcolor{black}{0.289} & \textcolor{black}{0.266} & \textcolor{black}{0.320} & \textcolor{black}{0.352*} & \textcolor{black}{0.325} & \textcolor{black}{0.311} & \textcolor{black}{\textbf{0.536}} & \textcolor{black}{0.486} & \textcolor{black}{52.3\%} \\
\bottomrule
\end{tabular}
\end{table*}

\subsection{Experimental setting}

\subsubsection{Metrics} We assess model performance from three perspectives. For node classification, we use accuracy following prior works \citep{DCIWang2021decoupling,DLRGAEchen2023dual,SGCwu2019simplifying}. To evaluate explanation quality, we employ the faithfulness metric as described in Definition \ref{Definition 2}. Model training efficiency is measured by the computational duration.

\subsubsection{Baselines} We compare BAED against ten baselines in classification performance. \textcolor{black}{SAGE} \citep{hamilton2017inductive}, GAT \citep{velivckovic2017graph}, GIN \citep{GIN}, and GCN \citep{kipf2016semi} represent traditional GNNs without specific designs for few-shot scenarios. 
While, SGC \citep{SGCwu2019simplifying}, DLRGAE \citep{DLRGAEchen2023dual},
\textcolor{black}{and HiD-Net \citep{li2024hidnet} for more advanced GNNs, as well as GPN \citep{ding2020gpn},
tsGCN \citep{tsGCNwu2023}, and DCI \citep{DCIWang2021decoupling} standing for SOTA baselines tailored for FSGL.} 
For evaluating interpretability, GE-L \citep{GE-L}, LIME \citep{LIME}, DeepWalk (DW) \citep{DW}, LN-L \citep{LN-L}, and LP \citep{LP} are the control group of non-GNN-based explanation algorithms. They are compared with seven gradient-based methods under the BAED framework: SM \citep{SM}, inpgrad \citep{inpgrad}, IG \citep{IG}, pgexp \citep{pgexp}, gnnexp \citep{gnnexp}, gbp \citep{gbp}, and deconv \citep{deconv}. Detailed descriptions of all baselines are provided in \textsc{Appendix} \textsc{\ref{Appendix 2}}.

\subsubsection{Implementation} The hyperparameters of baselines are set by default as the original papers. In the main body, the labeling ratio is set at 1\%, unless otherwise specified in sensitivity analysis. To support models requiring feature embeddings, the features for unlabeled nodes are initialed with the average of labeled nodes' features. The default backbone of auxiliary GNN in BAED is \textcolor{black}{SAGE} \citep{hamilton2017inductive}, and the subgraph extraction method is IG \citep{IG}, denoted as ``\textcolor{black}{SAGE+IG}'' in Table~\ref{tabClassification}. For a fair comparison, all the methods share the same training parameters. More details are provided in \textsc{Appendix} \textsc{\ref{Appendix imp}}. 

\subsection{Overall Performance on Classification}

\textcolor{black}{Table \ref{tabClassification} lists} 
the classification accuracy of all baselines, with bold font highlighting the better performance using the BAED framework, and the numbers with an asterisk indicate the best performance among baselines. \textbf{Overall, in feature-agnostic FSGL scenarios, BAED improves \textcolor{black}{classification} effectiveness by an average of \textcolor{black}{50.2\%} across \textcolor{black}{eight} datasets from different domains. }

We further analyze some noteworthy phenomena: 
First, comparing the traditional GNNs with BAED, the core difference lies in whether the explanation-in-the-loop technique is incorporated. Basic GNNs directly output class probabilities, while in BAED, GNNs serve as auxiliary means for supporting BP in predicting on the explanatory subgraph. \textbf{The advanced effectiveness demonstrates the better adaptability of explanation-based techniques for FSGL problems than direct prediction with GNNs.}

Second, we observe that existing FSGL models do not significantly outperform traditional GNNs. A potential reason is that current FSGL models are essentially based on the similarity between the labeled node features and the target node features. In more general feature-agnostic cases with only graph structure and labeled nodes' information, FSGL models fail to map the relationship between features and labels. In contrast, \textbf{BAED inherently learns the mapping between prior and posterior probabilities to identify the topological relationships between pseudo-labels, thus reducing dependence on node features.}

Third, BAED does not outperform existing methods on CoauthorCS, which is \textcolor{black}{a} very dense graph, with 15 classes and an average degree of 18. In such case, BAED's advantages in handling sparse information diminish because performing BP on such a dense graph leads to messages transmitted to the target node containing interfering label information from different classes.

\textcolor{black}{We adopt a 1\% labelling ratio to better demonstrate the advantages of BAED under varying numbers of shots, which closely aligns with the conventional few-shot setting. 
For example: The Wiki dataset has 2,405 nodes and 17 classes, yielding an average of $2,405/17*0.01  \approx  1.4$ shots per class, 
and similarly Wisconsin dataset leads to about 1 ($251*0.01/5\approx0.5$) shot case; 
The Cora dataset has 2,708 nodes and 7 classes, giving an average of $2708/7*0.01  \approx 3.87$ shots per class; The larger CoauthorCS dataset has 18,333 nodes and 15 classes, with $18,333/15*0.01  \approx  12.2$ shots per class, which is close to a 10-shot setting. Even under these varying shot scenarios, 
\textcolor{black}{BAED consistently maintains robust performance in terms of accuracy, efficiency, and explainability across experiments.}
According to the definition of model robustness, namely the ability to maintain stable performance under different levels of attack and abnormal situations \citep{survey-attack}, reporting results in terms of labelling ratio provides a more comprehensive and rigorous evaluation.}

\begin{table*}[tbhp]\footnotesize
\centering
\renewcommand{\arraystretch}{1.12}
\setlength\tabcolsep{4.6 pt}
\caption{Overall performance comparison on explainability \textbf{$\downarrow$}}
\label{tabExp}
\begin{tabular}{@{}l||ccccccc|ccccc|c@{}}
\toprule
\multirow{2}{*}{Datesets} &
  \multicolumn{7}{c|}{\textbf{BAED-enabled} gradient-based explanation models} &
  \multicolumn{5}{c|}{Non-GNN-based explanation algorithms} &
  \multirow{2}{*}{Improve.} \\ \cmidrule(lr){2-13}
 &
  \multicolumn{1}{c|}{SM} &
  \multicolumn{1}{c|}{inpgrad} &
  \multicolumn{1}{c|}{IG} &
  \multicolumn{1}{c|}{pgexp} &
  \multicolumn{1}{c|}{gnnexp} &
  \multicolumn{1}{c|}{gbp} &
  deconv &
  \multicolumn{1}{c|}{GE-L} &
  \multicolumn{1}{c|}{LIME} &
  \multicolumn{1}{c|}{DW} &
  \multicolumn{1}{c|}{LN-L} &
  LP &
   \\ \midrule
Cora       & 0.016 & \textbf{0.010} & \textbf{0.010} & 0.018          & 0.018 & 0.014          & 0.015          & 0.015* & 0.060 & 0.017  & 0.017  & -      & 33.3\% \\
Citeseer   & 0.046 & 0.039          & 0.041          & 0.040          & 0.042 & \textbf{0.030} & 0.031          & 0.030* & 0.274 & 0.041  & 0.042  & 0.045  & 0.0\%  \\
Pubmed     & 0.273 & 0.264          & \textbf{0.258} & 0.393          & 0.289 & 0.268          & 0.261          & 0.287  & 0.611 & 0.384  & 0.376  & 0.266* & 3.0\%  \\
Wiki       & 1.178 & 1.178          & 1.164          & 1.179          & 1.177 & 1.157          & \textbf{1.139} & 1.149* & 1.489 & 1.180  & 1.180  & -      & 0.9\%  \\
DBLP       & 1.878 & \textbf{1.799} & 1.825          & 2.615          & 2.050 & 1.851          & 1.848          & 2.135* & 3.600 & 2.612  & 2.634  & 2.192  & 15.7\% \\
Wisconsin  & 0.007 & 0.007          & 0.007          & \textbf{0.006} & 0.007 & 0.007          & 0.007          & 0.007* & 0.067 & 0.007* & 0.007* & 0.008  & 14.3\% \\
CoauthorCS & 3.161 & \textbf{3.143} & 3.167          & 3.952          & 3.521 & 3.199          & 3.201          & 3.595* & 4.541 & 3.953  & 3.935  & 3.720  & 12.6\% \\ 
\textcolor{black}{CoauthorPhy} & \textcolor{black}{5.331} & \textcolor{black}{\textbf{5.115}} & \textcolor{black}{5.163} & \textcolor{black}{8.543} & \textcolor{black}{6.650} & \textcolor{black}{5.302} & \textcolor{black}{5.295} & \textcolor{black}{6.530*} & \textcolor{black}{12.672} & \textcolor{black}{8.833} & \textcolor{black}{8.761} & \textcolor{black}{7.276} & \textcolor{black}{21.7\%} \\
\bottomrule
\end{tabular}
\end{table*}

\begin{table*}[tbhp]\footnotesize
\centering
\renewcommand{\arraystretch}{1.12}
\setlength\tabcolsep{3.5 pt}
\caption{Overall performance comparison on efficiency}
\label{tabEff}
\begin{tabular}{@{}l||ccccc||c|ccccc|c@{}}
\toprule
\multirow{2}{*}{Datasets} &
  \multicolumn{5}{c||}{Number of epochs per second \textbf{$\uparrow$}} &
  \multirow{2}{*}{\begin{tabular}[c]{@{}c@{}}BP time in \\ BAED (sec.)\end{tabular}} &
  \multicolumn{5}{c|}{Total training time in seconds \textbf{$\downarrow$}} &
  \multirow{2}{*}{Improve.} \\ \cmidrule(lr){2-6} \cmidrule(lr){8-12}
 &
  \multicolumn{1}{c|}{\textbf{BAED}} &
  \multicolumn{1}{c|}{GCN} &
  \multicolumn{1}{c|}{\textcolor{black}{SAGE}} &
  \multicolumn{1}{c|}{tsGCN} &
  DCI &
   &
  \multicolumn{1}{c|}{\textbf{BAED}} &
  \multicolumn{1}{c|}{GCN} &
  \multicolumn{1}{c|}{\textcolor{black}{SAGE}} &
  \multicolumn{1}{c|}{tsGCN} &
  DCI & 
   \\ \midrule
Cora       & \textbf{19.46}   & 4.63 & 1.88          & 2.81 & 5.03          & 1.13 & \textbf{6.1}  & 58.3  & 76.5          & 198.0 & 29.1  & 78.9\%  \\
Citeseer   & \textbf{10.10}   & 5.47 & 3.40          & 3.28 & 3.81          & 0.11 & \textbf{10.1} & 31.9  & 30.3          & 159.2 & 32.0  & 66.3\%  \\
PubMed     & 2.60             & 2.21 & 1.66          & 0.84 & \textbf{2.82} & 0.06 & \textbf{38.1} & 51.6  & 60.0          & 827.0 & 653.1 & 26.8\%  \\
Wiki       & 3.65             & 3.43 & 2.21          & 1.98 & \textbf{5.03} & 0.15 & \textbf{27.1} & 29.4  & 45.2          & 268.1 & 32.4  & 6.4\%   \\
DBLP       & 1.67             & 1.94 & \textbf{2.60} & 1.00 & 1.19          & 1.11 & 61.1          & 50.8  & \textbf{38.7} & 716.8 & 218.9 & -56.7\% \\
Wisconsin  & \textbf{1446.67} & 4.51 & 8.04          & 7.56 & 7.00             & 0.01 & \textbf{2.0}  & 23.0    & 12.1          & 66.6  & 6.0   & 66.6\%  \\
CoauthorCS & \textbf{3.62}    & 1.00 & 0.41          & 0.88 & 1.13          & 1.73 & \textbf{78.7} & 521.1 & 558.6         & 757.3 & 545.2 & 84.9\%  \\ 
\textcolor{black}{CoauthorPhy} & \textcolor{black}{\textbf{2.43}} & \textcolor{black}{0.65} & \textcolor{black}{0.14} & \textcolor{black}{0.12} & \textcolor{black}{1.34} & \textcolor{black}{3.65} & \textcolor{black}{\textbf{44.7}} & \textcolor{black}{1173.4} & \textcolor{black}{1732.8} & \textcolor{black}{2291.0} & \textcolor{black}{1574.9} &  \textcolor{black}{96.2\%} \\
\bottomrule
\end{tabular}
\end{table*}

\subsection{Overall Performance on Explainability}
\label{sec-exp}

High-quality explanations are crucial for an explanation-in-the-loop predictive model. In this section, we examine model performance from the perspective of explainability based on the metric faithfulness. BAED, aiming at a pioneering framework addressing the FSGL problem, enables a majority of gradient-based explanation models, which are categorized on the left sector of Table~\ref{tabExp}. For the control group, we utilize non-GNN-based explanation algorithms that do not require the parameter training phase, such as LIME and DeepWalk. 

Overall, the BAED-enabled category demonstrates higher faithfulness in explanations, validating the broad applicability of BAED in addressing explainability issues in FSGL. \textbf{The average faithfulness of the BAED-enabled category across \textcolor{black}{eight} datasets exceeds that of non-GNN-based algorithms by \textcolor{black}{12.7}\%.} This improvement indicates that leveraging label augmentation through BP and auxiliary GNNs can help these gradient-based models better identify explanatory information, thereby supporting the final predictions. 

\subsection{Model Efficiency Analysis}

In this section, we compare BAED with baselines in terms of model efficiency. As shown in Table~\ref{tabEff}, 
\textcolor{black}{BAED demonstrates clear advantages along two dimensions.}
The number of epochs per second indicates that BAED achieves faster data throughput when processing the same volume of samples. Total training time also confirms the higher convergence speed of BAED for training the parameters. This advantage arises from two main factors. First, the approach to handling raw features is different. Traditional GNNs feed long feature vectors directly into a neural network, which are then processed through several layers of MLP before entering the GNN aggregation stage. In contrast, BAED employs a rapid BP process to first convert raw features into prior probabilities, the length of which is equal to the number of classes. As illustrated in dataset statistics, the number of classes is far smaller than the number of features. Thus, \textbf{BAED allows for a substantial reduction in the number of trainable parameters required, specifically from the input layer to the aggregation layer of the GNNs.} Second, the computation time required for the BP process until its convergence is negligible compared to the training time of the neural network, averaging only 4.2\% of the total training time. This number demonstrates that \textbf{the BP algorithm, within the FSGL problem, serves not only as a means of label augmentation but also as an efficient feature preprocessing method that enhances model efficiency.} 
\textcolor{black}{In addition, BAED’s convergence speed is limited on DBLP, where the input and output vector lengths are only three, despite the graph’s high density (average degree of 12).}
The small number of trainable parameters in the auxiliary GNN requires more iterations to capture the distinguishable characteristics between nodes.

\subsection{Ablation Study}
\label{abla}

Due to the modular design of BAED, we can remove each module to create three ablation models from BAED: Ablation-1 uses BP on raw graphs to compute posterior probabilities as predictions; Ablation-2 uses label-augmented graphs to train GNNs and make predictions; and Ablation-3 uses raw graphs to train GNNs and make predictions. Their performances compared to the full BAED framework are shown in Table~\ref{Tab_Ablation}. 

\begin{table}[bthp] \footnotesize
\centering
\renewcommand{\arraystretch}{1.3}
\setlength\tabcolsep{4 pt}
\caption{Classification Accuracy (\%) of Ablation Models }
\begin{tabular}{@{}l||ccc|c@{}}
\toprule
\multirow{2}{*}{Datasets} & \multicolumn{1}{l|}{Ablation-1} & \multicolumn{1}{l|}{Ablation-2} & Ablation-3                & \multirow{2}{*}{\textbf{BAED}} \\ \cmidrule(lr){2-4}
                          & \multicolumn{1}{c|}{\textbf{B}\,\st{AED}}       & \multicolumn{1}{c|}{\textbf{BA}\,\st{ED}}       & \multicolumn{1}{c|}{\st{B}\,\textbf{A}\,\st{ED}} &                       \\ \midrule
Cora       & 16.70          & 30.15          & 30.15          & \textbf{51.81} \\
Citeseer   & 8.97           & 18.16          & 20.21          & \textbf{78.67} \\
PubMed     & 47.46          & 39.39          & 40.00          & \textbf{50.67} \\
Wiki       & 8.10           & \textbf{16.12} & 15.72 & 12.71          \\
DBLP       & 55.09          & 44.26          & 44.26          & \textbf{69.75} \\
Wisconsin  & 35.80          & 26.75          & 26.75          & \textbf{54.73} \\
CoauthorCS & \textbf{45.18} & 12.60          & 22.57          & 19.24          \\ \bottomrule
\end{tabular}
\label{Tab_Ablation}
\end{table}

Here are some key conclusions: First, the main difference between Ablation-1 and the full BAED pipeline is that Ablation-1 predicts on the raw graph with BP, whereas BAED makes decisions on the explanatory subgraph. The classification performance of Ablation-1 decreases by an average of $17$ percentage points. As stated in the motivation, although BP can capture information across a broader scope, the sparsity of labels on the original graph makes BP vulnerable to excessive noise in the neighborhood. In contrast, the auxiliary GNNs and explanatory subgraph extraction modules in BAED resolve this bottleneck.

Second, neither Ablation-2 nor Ablation-3 adopts the technique of explanation in the loop. Compared to the full BAED, their accuracy declines by $21$ and $20$ percentage points, respectively. This comparison validates the fundamental concept of explanation-guided learning in BAED, demonstrating that high-quality explanations can further provide positive feedback and enhance model performance in downstream tasks.

\begin{figure*}[t]
\centering
\includegraphics[width=6.2 in]{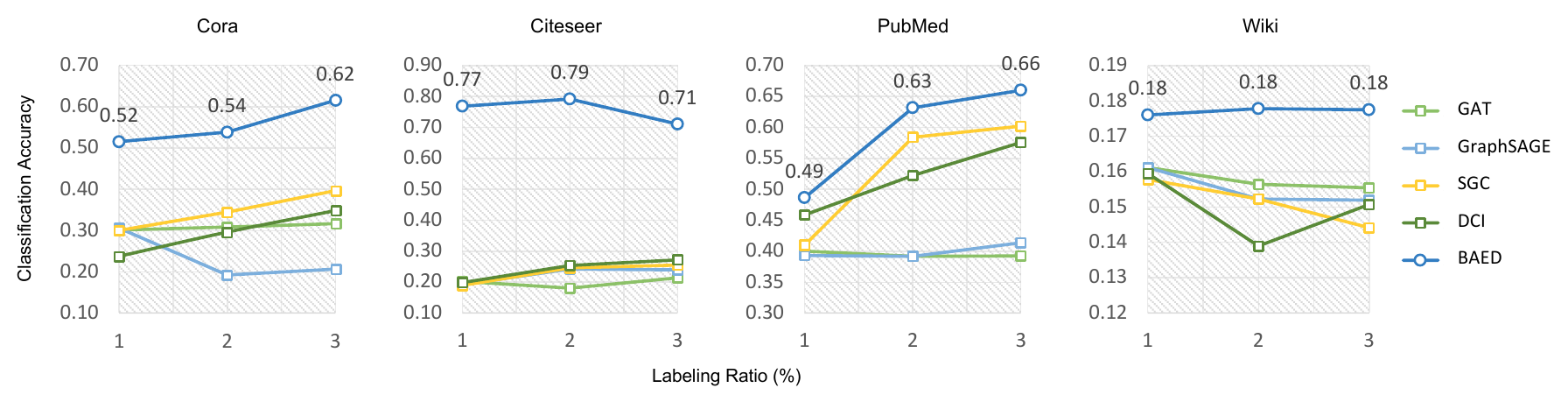}
\caption{\textcolor{black}{Classification accuracy with different labeling ratios}}
\label{Fig_Ratio}
\end{figure*}

Third, the primary difference between Ablation-2 and Ablation-3 is that Ablation-2 uses the prior probability vectors on the augmented graph as the input for GNNs, while Ablation-3 directly uses the raw feature vectors as input. Ablation-3 only outperforms Ablation-2 by 1.8 percentage points. This aligns with our analysis in the efficiency section: compared to directly processing raw features with GNNs, using BP for label augmentation can preserve important label information and significantly reduces model training time, improving training speed by nearly tenfold.

\subsection{Sensitivity Analysis on Labeling Ratios}
\label{sec:sensitivity_analysis}

\textcolor{black}{We evaluated BAED against major baselines under various labeling ratios.}
Figure~\ref{Fig_Ratio} visually demonstrates that BAED consistently outperforms basic GNNs and existing FSGL methods across various datasets and labeling ratios. Overall, as the labeling ratio increases, the performance of BAED and the other models also improves. Additionally, when summarizing results from the dimension of labeling ratios, we find that averaging on seven datasets, BAED outperforms the second-best baseline by 50\%, 39\%, and 37\% when the labeling ratios are 1\%, 2\%, and 3\%, respectively. \textbf{This indicates that BAED's effectiveness is more pronounced in graphs with sparser labels.} For specific reasons, please refer to the analysis in the overall classification performance section, which discusses the impact of information density on FSGL models and BAED. \textcolor{black}{Due to space constraints, we leave the experiment results about other three datasets and the impact of labeling ratios on model interpretability in the \textsc{Appendix} \textsc{\ref{Appendix exp}}.}

\subsection{Sensitivity Analysis on the Sizes of Explanatory Subgraph}
\label{sec:sensitivity_analysis_size}

To further investigate how different number of nodes in the explanatory subgraph will affect the model’s performance. We perform sensitivity analysis as shown in Table~\ref{Tab_Size}. As the subgraph size increases, accuracy (\%) rapidly improves and then gradually levels off. When the graph size exceeds a certain value, larger subgraphs do not provide more discriminative but noisy information.

\begin{table}[bthp] \footnotesize
\centering
\renewcommand{\arraystretch}{1.3}
\setlength\tabcolsep{4 pt}
\caption{Classification Accuracy (\%) with Various Sizes of Explanatory Subgraphs }

\begin{tabular}{@{}l||ccccccc@{}}
\toprule
Size of Explanatory Subgraph        & 2    & 3    & 4    & 5    & 6    & 7    & 8    \\ \midrule
Cora       & 50.4 & 50.8 & 51.1 & 51.8 & 51.7 & 49.9 & 49.7 \\
Citeseer   & 77.3 & 78.1 & 78.1 & 78.7 & 78.8 & 79.5 & 80.0 \\
Pubmed     & 46.7 & 48.3 & 49.7 & 50.7 & 51.0 & 51.3 & 51.5 \\
Wiki       & 12.1 & 12.3 & 12.5 & 12.7 & 12.9 & 13.1 & 13.3 \\
dblp       & 68.9 & 68.9 & 69.8 & 69.8 & 70.1 & 70.1 & 70.1 \\
Wisconsin  & 26.1 & 27.8 & 29.8 & 42.0 & 51.8 & 51.0 & 32.7 \\
coauthorCS & 18.4 & 41.0 & 16.6 & 19.2 & 19.5 & 19.5 & 19.9 \\ \bottomrule
\end{tabular}

\label{Tab_Size}
\end{table}

\begin{table}[!hbp]
\centering
\caption{\textcolor{black}{Prediction (accuracy) and explanation (faithfulness) performance of BAED under various $\epsilon$ on Cora dataset.} }
\textcolor{black}{
\begin{tabular}{c||cccccccccc}
    \toprule
    $\epsilon$ & 0.60 & 0.65 & 0.70 & 0.75 & 0.80 & 0.85 & 0.90 & 0.95 & 1.00 \\
    \midrule
    accuracy $\uparrow$ & 0.549 & 0.565 & 0.582 & 0.517 & 0.500 & 0.529 & 0.518 & 0.483 & 0.582 \\
    faithfulness $\downarrow$ & 0.004 & 0.004 & 0.004 & 0.004 & 0.004 & 0.005 & 0.016 & 0.009 & 0.010 \\
    \bottomrule
\end{tabular}
}
    \label{tab:sensitivity_epsilon}
\end{table}

\textcolor{black}{
\subsection{Sensitivity Analysis on Compatibility Hyperparameter}
To study the impact of compatibility hyperparameter, e.g., $\epsilon$ in equation \ref{eq_comp}, Table \ref{tab:sensitivity_epsilon} shows BAED's prediction and explanation performance when $\epsilon$ is set from 0.60 to 1.00. 
BAED exhibits stable and superior performance under different settings.
More importantly, comparing to the results in Tables \ref{tabClassification} and \ref{tabExp}, 
BAED consistently outperforms baselines significantly as $\epsilon$ changes.
}

\subsection{Convergence Test of BP on Cyclic Graph}

As an iterative method, BP can be theoretically proven to converge on acyclic graphs \citep{bp1}. As a core technique in BAED, it is essential to verify that it meets the convergence-stopping criteria on large-scale cyclic graphs; otherwise, the subsequent operations will be unfeasible. Here, we plot the convergence curves of the BP process in the first label augmentation module. Generally, all eight datasets under different labeling ratios can reach a strict convergent state within $20$ iterations. The convergence tests for other labeling ratios are shown in the \textsc{Appendix} \textsc{\ref{Appendix cvg}}. In addition, the BP process on explanatory subgraphs is guaranteed theoretically and empirically, since the vast majority of subgraphs with $5$ nodes are acyclic \citep{GE-L}.

\begin{figure}[tbhp]
\centering
\includegraphics[width=2.8 in]{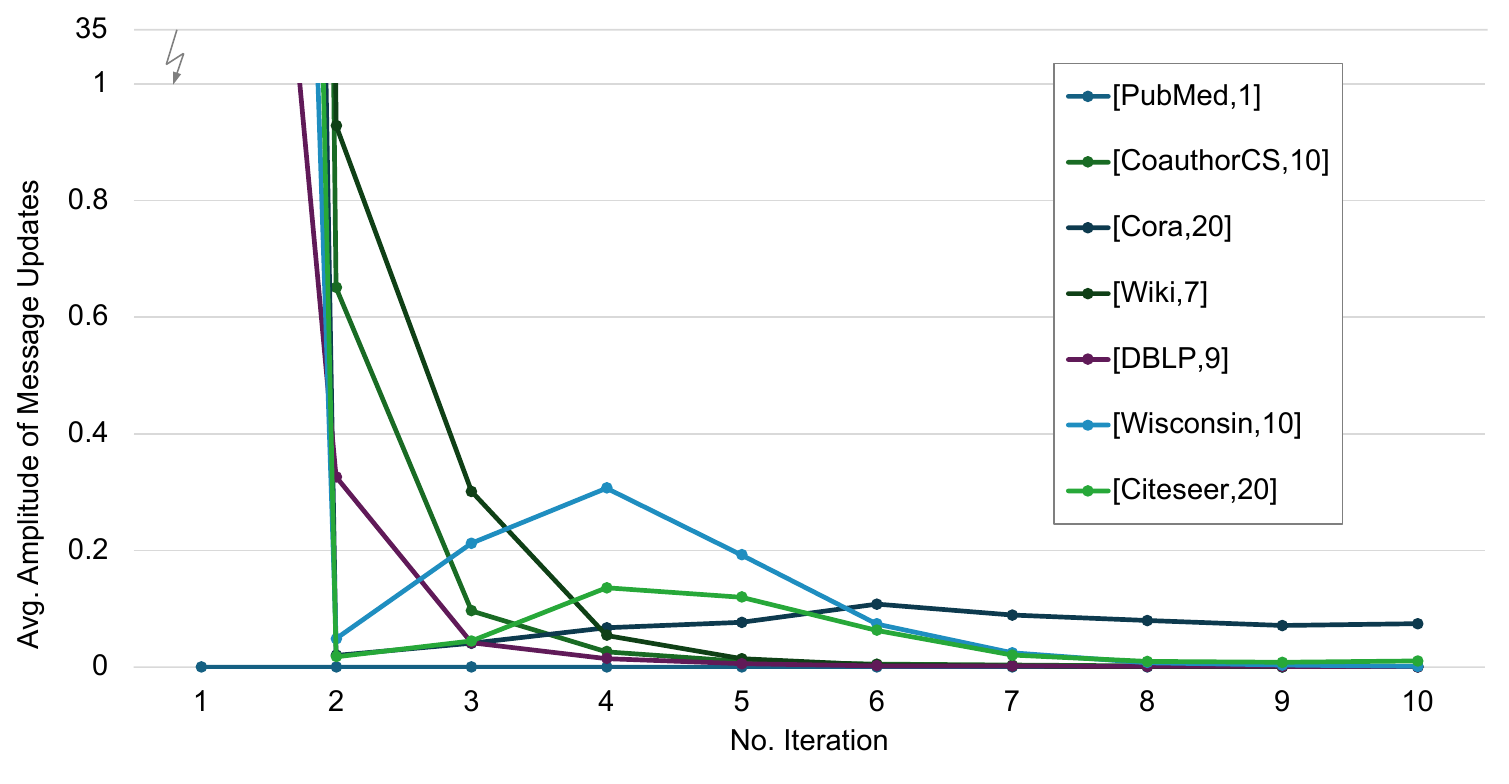}
\caption{Convergence Curves of BP Process in Label Augmentation. Legend denotes [dataset, No. converged iteration]}
\label{Fig_Convg}
\end{figure}


\section{Conclusion and Future Work}

In this work, 
\textcolor{black}{BAED achieves comprehensive improvements in effectiveness, efficiency, and explainability for FSGL}
through innovative BP-based label augmentation, subgraph extraction techniques, and explanation-in-the-loop prediction. As a new paradigm and fundamental framework for explainable FSGL, BAED maintains high compatibility, allowing for further exploration within research areas such as graph augmentation technique, GNN backbone design, explanation methods, and explanation-guided learning. 
\textcolor{black}{
\textcolor{black}{Some future directions}
are outlined based on the capacity of BAED and findings in our experiments: First, BAED is not limited to gradient-based methods and can be adapted to various subgraph extraction approaches. As demonstrated in our experiments, the quality of the explanatory subgraphs is highly correlated with the final prediction accuracy. Introducing more subgraph extraction algorithms, such as large language model-driven explanation methods, represents a promising future direction. Second, the BAED pipeline can be further extended to other tasks such as edge classification, graph classification, and temporal graph prediction. These scenarios provide opportunities for explanation-guided learning research in broader contexts. Third, applying BAED to real-world scenarios, such as anti-money laundering and social network fraud detection, is another valuable direction.}


\clearpage
\bibliography{main.bib}
\bibliographystyle{model1-num-names}

\appendix

\section{Appendix}

\subsection{Baselines}
\label{Appendix 2}
\subsubsection{GNNs for Classification}
\begin{itemize}[leftmargin=1em]
 \item \textcolor{black}{SAGE} \citep{hamilton2017inductive} generates node embeddings by aggregating features from local neighborhoods, allowing nodes to efficiently handle previously unseen ones.

 \item GAT \citep{velivckovic2017graph} introduces a novel architecture that utilizes attention to improve GNNs, learning edge weights without complex operations or prior graph knowledge.

 \item GIN \citep{GIN} uses a sum-based aggregation mechanism, where the node embedding is updated by aggregating the features from neighbors with a non-linear layer.

 \item GCN \citep{kipf2016semi} provides a novel paradigm for the learning on graphs by applying convolution operations directly to graphs via the first-order approximation.
\end{itemize}

\subsubsection{FSGL models for Classification}
\begin{itemize}[leftmargin=1em]
 \item SGC \citep{SGCwu2019simplifying} eliminates non-linearities and collapsing weights between consecutive layers, resulting in a simplified linear model that reduces the complexity of traditional GCNs.

 \item DLRGAE \citep{DLRGAEchen2023dual} improves on traditional graph autoencoders by incorporating semantic and topological homophily to capture the relationships between nodes in two distinct graphs. Explores low-rank information across both semantics and topology.

 \item tsGCN \citep{tsGCNwu2023} introduces an interpretable optimization framework that enables the design and interpretation of various GCN models by incorporating the appropriate regularizers.

 \item DCI \citep{DCIWang2021decoupling} is a graph self-supervised learning (SSL) scheme designed for node representation learning, which improves anomaly detection by clustering the graph into multiple parts to capture intrinsic properties in concentrated feature spaces. 
\end{itemize}

\subsubsection{Gradient-based models for Explanation}
\begin{itemize}[leftmargin=1em]
 \item SM \citep{SM} is a visualization technique for deep convolutional networks, where the gradient of the class score is computed to highlight the most important areas that contribute to a specific class prediction.

 \item inpgrad \citep{inpgrad} decomposes the prediction from models via searching the contribution score of each neuron to the features, providing a clear interpretability of decision making.

 \item IG \citep{IG} attributes a network's prediction to the input from the gradients of the class score, offering a straight approach for the interpretability of models.

 \item pgexp \citep{pgexp} is a parameterized explainer designed to solve the limitations in focusing on local, instance-specific explanations, enabling multi-instance explanations, and offering better generalization and applicability in inductive settings. 

 \item gnnexp \citep{gnnexp} identifies key subgraph structures and relevant node features that significantly influence prediction. It maximizes mutual information between the prediction and possible subgraphs, generating consistent and concise explanations across instances.

 \item gbp \citep{gbp} is an approach that observes the changes in prediction with perturbation. This approach identifies what most influences the network's decision, providing insights into the network's internal mechanisms.

 \item deconv \citep{deconv} is a framework that learns robust low and mid-level representations by capturing complex cues. It involves the unsupervised convolutional decomposition under a sparsity constraint, enabling features to emerge spontaneously from the data.
\end{itemize}

\subsubsection{Non-GNNs-based algorithms for Explanation}
\begin{itemize}[leftmargin=1em]
 \item GE-L \citep{GE-L} is an explanation method designed to prioritize speed over explanation fidelity with a beam search in subgraph space.

 \item LIME \citep{LIME} is designed to provide interpretable and faithful explanations for any classifier's predictions with a simpler model. It frames the task of presenting non-redundant explanations for individual predictions as a submodular optimization problem.

 \item DW \citep{DW} learns latent representations of nodes by treating random walks as sequences. It encodes social relationships from local information within a continuous space.

 \item LN-L \citep{LN-L} is designed for embedding sophisticated models into simplified vector spaces. It efficiently handles networks by optimizing an objective that utilizes both local and global information by edge-sampling, improving effectiveness and efficiency.

 \item LP \citep{LP} is a semi-supervised approach that represents (un)labeled data as vertices in a graph. It incorporates class priors and predictions from supervised classifiers and proposes a method for parameter learning via entropy minimization.
\end{itemize}

\subsection{Implementation Details} 
\label{Appendix imp}

\subsubsection{Data Prepossessing} The datasets are built and imported from an open-source repository, \texttt{torch\_geometric}. The statistical details are shown in Table \ref{tab:datasets}. To train and evaluate the model performance in various experimental settings, the training portion consists of only a few nodes ranging from $1\%$, $2\%$, and $3\%$. $200$ of the unlabeled nodes will be randomly selected as the number of target nodes.

\subsubsection{Model Implementation} The training phase will stop the strategy if the models do not achieve better performance within 1000 epochs. The hidden dimension is 32, and the learning rate is 0.1. As for the details of BAED, $\epsilon$ in Eq. (\ref{eq_comp}) is 0.1. Belief Propagation for both steps will be terminated if the messages converge with average changes smaller than $\eta$= 0.001 or the BP runs out of 20 iterations. The primary experiments are run on the Macbook Pro without GPU and use Pytorch 1.4 by default. Research indicates that humans can typically handle an average of ($7 \pm 2$) entities simultaneously \citep{miller1956}. Therefore, we limit the size of the explanatory subgraphs to a maximum of five nodes (C=5) in order to achieve a balance between simplicity and faithfulness.

\subsection{Classification performance with different labeling ratios} 
\label{Appendix class}

\begin{figure*}[thbp]
\centering
\includegraphics[width=6.2 in]{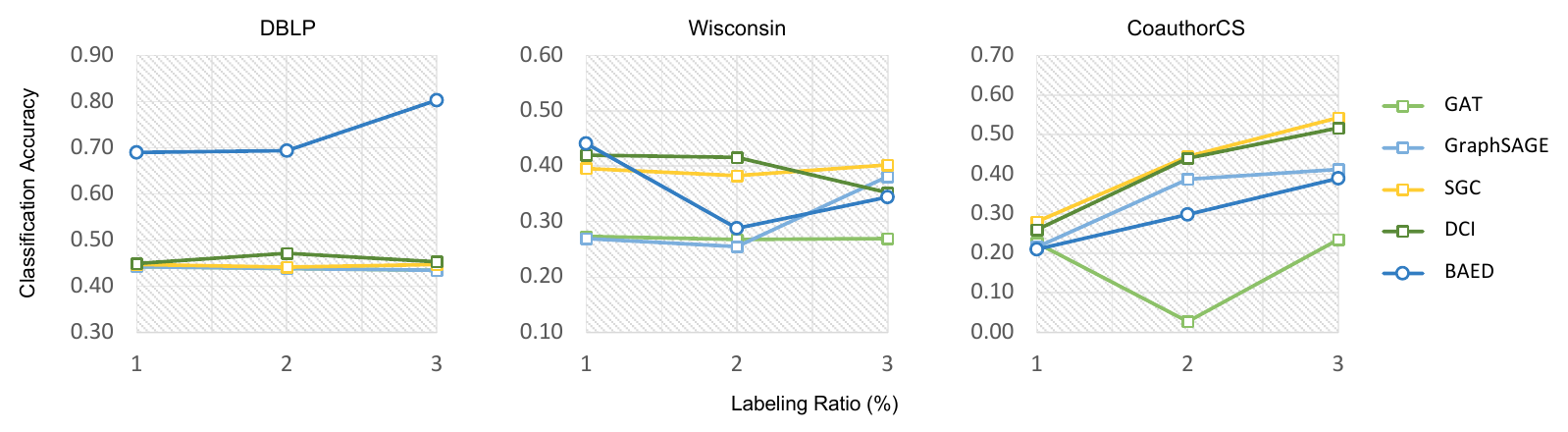}
\caption{\textcolor{black}{Classification accuracy with different labeling ratios}}
\label{Fig_Ratio2}
\end{figure*}

In Table \ref{table-class-append}, we list the classification performance of all baselines with different labeling ratios. And Figure \ref{Fig_Ratio2} plots the corresponding results for DBLP, Wisconsin, and CoauthorCS datasets.

\begin{table*}[t]\small
\centering
\renewcommand{\arraystretch}{1.2}
\setlength\tabcolsep{1 pt}
\caption{Classification performance with different labeling ratios}
\label{table-class-append}
\begin{tabular}{@{}l||c|cccc|cccc|cc}
\toprule
\multirow{2}{*}{Datasets} & \multirow{2}{*}{ratio(\%)} &
  \multicolumn{4}{c|}{GNNs} &
  \multicolumn{4}{c|}{FSGL Models} &
  \multicolumn{2}{c}{BAED} \\
\cmidrule(lr){3-12}
& & 
  \multicolumn{1}{c|}{\textcolor{black}{SAGE}} &
  \multicolumn{1}{c|}{GAT} &
  \multicolumn{1}{c|}{GIN} &
  GCN &
  \multicolumn{1}{c|}{SGC} &
  \multicolumn{1}{c|}{DLRGAE} &
  \multicolumn{1}{c|}{tsGCN} &
  DCI &
  \multicolumn{1}{c|}{\textcolor{black}{SAGE+IG}} &
  \textcolor{black}{SAGE+SM}
  \\ \midrule

\multirow{3}{*}{Cora}   
& 2 & 0.309 & 0.193 & 0.120 & 0.348* & 0.345 & 0.303 & 0.237 & 0.296 & 0.539 & \textbf{0.603} \\
& 3 & 0.317 & 0.207 & 0.284 & 0.393 & 0.397* & 0.285 & 0.282 & 0.348 & \textbf{0.616} & 0.546 \\
\midrule

\multirow{3}{*}{Citeseer}   
& 2 & 0.180 & 0.242 & 0.219 & 0.241 & 0.247 & 0.221 & 0.273* & 0.254 & \textbf{0.792} & 0.779 \\
& 3 & 0.213 & 0.240 & 0.266 & 0.246 & 0.255 & 0.199 & 0.267 & 0.272* & 0.712 & \textbf{0.732} \\
\midrule

\multirow{3}{*}{PubMed}   
& 2 & 0.392 & 0.392 & 0.406 & 0.392 & 0.584* & 0.478 & 0.549 & 0.523 & \textbf{0.632} & 0.618 \\
& 3 & 0.392 & 0.414 & 0.435 & 0.392 & 0.602* & 0.540 & 0.581 & 0.576 & 0.660 & \textbf{0.663} \\
\midrule

\multirow{3}{*}{Wiki}     
& 2 & 0.161* & 0.161* & 0.161* & 0.161* & 0.158 & 0.161* & 0.159 & 0.141 & \textbf{0.141} &0.127\\
& 3 & 0.155* & 0.152 & 0.144 & 0.144 & 0.144 & 0.144 & 0.144 & 0.151 & \textbf{0.178} & 0.124 \\
\midrule

\multirow{3}{*}{DBLP}     
& 2 & 0.438 & 0.438 & 0.438 & 0.438 & 0.441 & 0.444 & 0.446 & 0.471* & 0.694 & \textbf{0.724} \\
& 3 & 0.434 & 0.434 & 0.434 & 0.434 & 0.447 & 0.437 & 0.447 & 0.453* & \textbf{0.803} & 0.703 \\
\midrule

\multirow{3}{*}{Wisconsin}  
& 2 & 0.267 & 0.255 & 0.267 & 0.263 & 0.383 & 0.272 & 0.267 & 0.416* & 0.288 & \textbf{0.547} \\
& 3 & 0.270 & 0.382 & 0.228 & 0.498* & 0.402 & 0.274 & 0.270 & 0.353 & 0.344 & \textbf{0.398} \\
\midrule

\multirow{3}{*}{CoauthorCS} 
& 2 & 0.027 & 0.387 & 0.227 & 0.267 & 0.445* & 0.156 & 0.300 & 0.440 & \textbf{0.298} & 0.274 \\
& 3 & 0.235 & 0.412 & 0.227 & 0.400 & 0.543* & 0.261 & 0.467 & 0.517 & 0.390 & \textbf{0.400}\\
\bottomrule
\end{tabular}
\end{table*}

\subsection{Explainability performance with different labeling ratios} 
\label{Appendix exp}

In Table \ref{table-exp-append}, we list the explainability evaluation of all baselines with different labeling ratios.

\begin{table*}[t]\small
\centering
\renewcommand{\arraystretch}{1.2}
\setlength\tabcolsep{2 pt}
\caption{Explainability performance with different labeling ratios}
\label{table-exp-append}
\begin{tabular}{@{}l||c|ccccccc|ccccc}
\toprule
\multirow{2}{*}{Datasets} & \multirow{2}{*}{Ratio (\%)} &
  \multicolumn{7}{c|}{BAED-compatible gradient-based explanation models} &
  \multicolumn{5}{c}{Non-GNN-based explanation algorithms} \\ \cmidrule(lr){3-14}
 & & SM & inpgrad & IG & pgexp & gnnexp & gbp & deconv & GE-L & LIME & DW & LN-L & LP \\
\midrule
\multirow{3}{*}{Cora}       
& 2  & \textbf{0.0200} & 0.0249 & 0.0246 & 0.0222 & 0.0220 & 0.0241 & 0.0238 & 0.0143* & 0.0710 & 0.0236 & 0.0226 & - \\
& 3  & 0.0164 & \textbf{0.0098} & 0.0099 & 0.0183 & 0.0180 & 0.0143 & 0.0155 & 0.0153* & 0.0601 & 0.0172 & 0.0174 & - \\
\midrule
\multirow{3}{*}{Citeseer}   
& 2  & 0.0145 & 0.0148 & \textbf{0.0130} & 0.0205 & 0.0174 & 0.0136 & 0.0136 & 0.0129* & 0.1466 & 0.0172 & 0.0176 & 0.0175 \\
& 3  & 0.0457 & 0.0392 & 0.0411 & 0.0403 & 0.0416 & \textbf{0.0304} & 0.0313 & 0.0302 & 0.2739* & 0.0405 & 0.0416 & 0.0453 \\
\midrule
\multirow{3}{*}{Pubmed}     
& 2  & 0.2512 & 0.2442 & 0.2507 & 0.3418 & 0.2497 & 0.2415 & \textbf{0.2332} & 0.2674 & 0.6142 & 0.3371 & 0.3307 & 0.2412* \\
& 3  & 0.2734 & 0.2643 & \textbf{0.2581} & 0.3927 & 0.2890 & 0.2685 & 0.2608 & 0.2868 & 0.6108 & 0.3841 & 0.3764 & 0.2656* \\
\midrule
\multirow{3}{*}{Wiki}       
& 2  & 1.1113 & 1.1143 & 1.1130 & \textbf{1.0943} & 1.1173 & 1.1143 & 1.1139 & 1.0835* & 1.2877 & 1.1226 & 1.1236 & - \\
& 3  & 1.1781 & 1.1781 & 1.1639 & 1.1787 & 1.1770 & 1.1574 & \textbf{1.1386} & 1.1485* & 1.4895 & 1.1797 & 1.1803 & - \\
\midrule
\multirow{3}{*}{DBLP}       
& 2  & \textbf{2.0111} & 2.0712 & 2.0651 & 2.2744 & 2.2239 & 2.0753 & 2.0647 & 2.1678 & 3.3435 & 2.4597 & 2.4608 & 2.1474* \\
& 3  & 1.8778 & \textbf{1.7988} & 1.8245 & 2.6148 & 2.0500 & 1.8513 & 1.8480 & 2.1348* & 3.5998 & 2.6125 & 2.6340 & 2.1919 \\
\midrule
\multirow{3}{*}{Wisconsin}  
& 2  & 0.0138 & 0.0206 & 0.0204 & \textbf{0.0106} & 0.0209 & 0.0219 & 0.0197 & 0.0173* & 0.0229 & 0.0189 & 0.0210 & 0.0217 \\
& 3  & 0.0069 & 0.0068 & 0.0072 & \textbf{0.0062} & 0.0066 & 0.0068 & 0.0066 & 0.0067* & 0.0675 & 0.0075 & 0.0071 & 0.0081 \\
\midrule
\multirow{3}{*}{CoauthorCS} 
& 2  & 3.1122 & \textbf{3.0925} & 3.1072 & 3.4579 & 3.2164 & 3.1178 & 3.1180 & 3.3208* & 3.8289 & 3.4937 & 3.4823 & 3.3691 \\
& 3  & 3.1606 & \textbf{3.1432} & 3.1666 & 3.9522 & 3.5206 & 3.1993 & 3.2006 & 3.5954* & 4.5408 & 3.9526 & 3.9348 & 3.7195 \\
\bottomrule
\end{tabular}
\end{table*}

\subsection{Convergence curves of BP process with different labeling ratios} 
\label{Appendix cvg}
\begin{figure*}[t]
\centering
\begin{minipage}{.49\textwidth}
    \includegraphics[width=3 in]{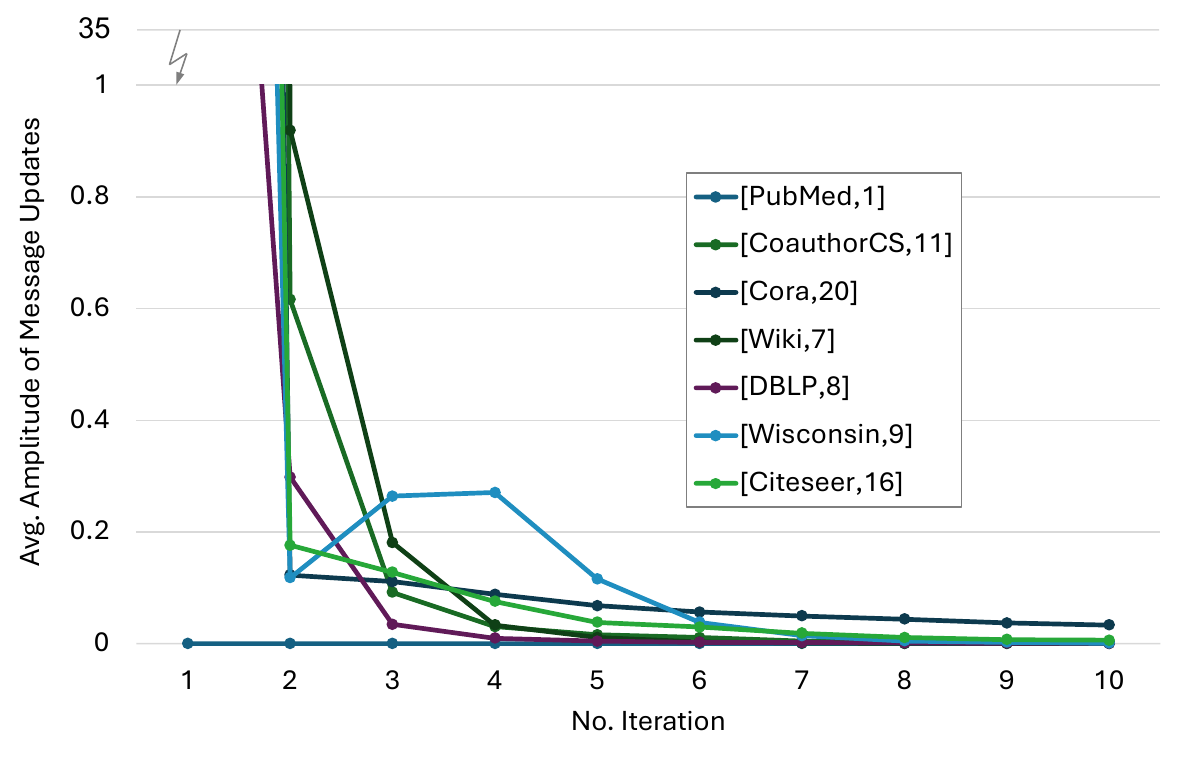}
\end{minipage}
\begin{minipage}{.49\textwidth}
    \includegraphics[width=3 in]{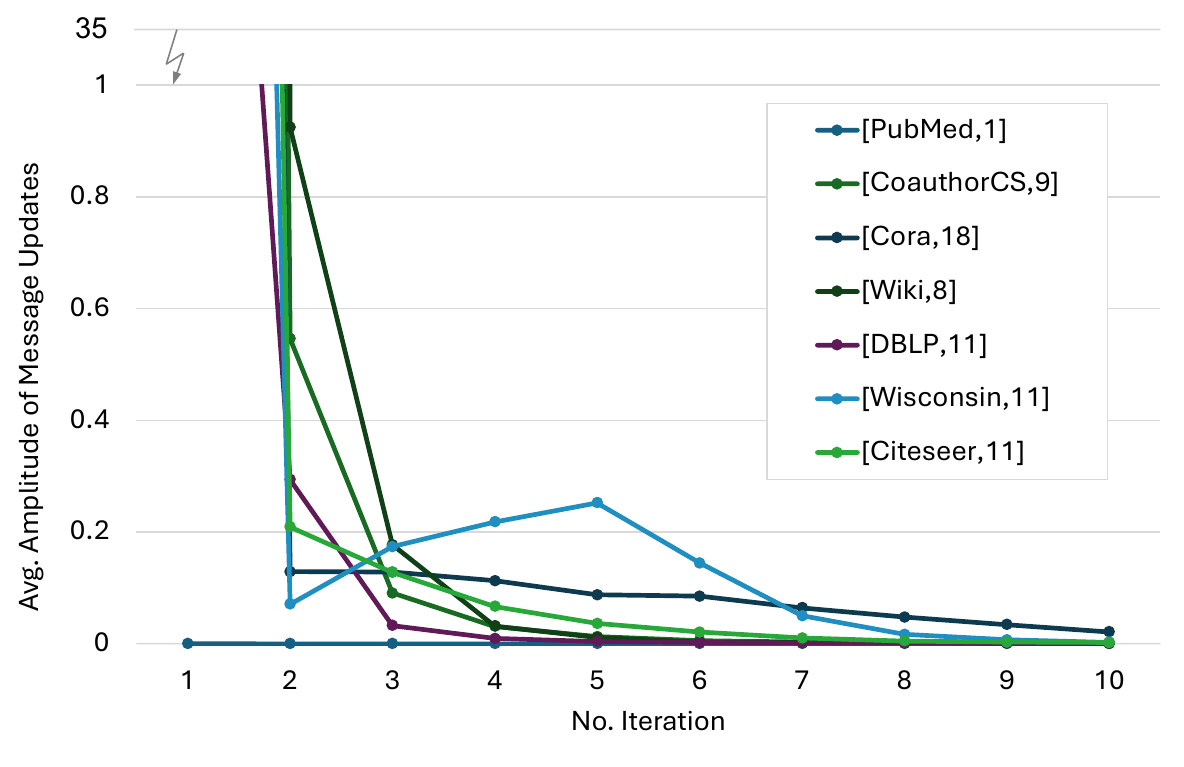}
\end{minipage}
    \caption{ 
        Convergence Curves of BP Process in Label Augmentation. Labeling Ratio = 2\% (Left) and 3\% (Right).
    }
    \label{fig_convg}
\end{figure*}

The convergence curves of BP process with different labeling ratios are shown in \textcolor{black}{Figure.} \ref{fig_convg}.



\end{document}